%% file: main.tex
\documentclass[runningheads]{llncs}

\usepackage{eccv}

\usepackage{eccvabbrv}
\usepackage{acronym}
\usepackage{soul}
\usepackage[ruled,vlined]{algorithm2e}
\usepackage{graphicx}
\usepackage{booktabs}

\usepackage[accsupp]{axessibility}  % Improves PDF readability for those with disabilities.

\usepackage{hyperref}

\usepackage{orcidlink}
\acrodef{ros}[ROS]{Robot Operating System}
\acrodef{apf}[APF]{Artificial Potential Field}
\acrodef{mpc}[MPC]{Model Predictive Controller}
\acrodef{at}[AT]{Assistive Technology}
\acrodef{bmi}[BMI]{brain-machine interface}

\usepackage{xcolor} % for color

\newcommand{\model}{OpenNav}

\usepackage{tikz}
\newcommand*\circled[1]{\tikz[baseline=(char.base)]{
            \node[shape=circle,draw,inner sep=2pt] (char) {#1};}}

\begin{document}

\title{\model: Efficient Open Vocabulary 3D Object Detection for Smart Wheelchair Navigation} 

\titlerunning{OpenNav}

\author{Muhammad Rameez ur Rahman\inst{1}\orcidlink{0000-0003-4425-5948} \and
Piero Simonetto\inst{2}\orcidlink{0009-0007-0164-3795} \and
Anna Polato\inst{2}\orcidlink{0000-0001-6348-6843}\and
Francesco Pasti\inst{2}\orcidlink{0009-0005-4992-3281}\and
Luca Tonin \inst{2}\orcidlink{0000-0002-9751-7190}\and
Sebastiano Vascon\inst{1,3}\orcidlink{0000-0002-7855-1641}}

\authorrunning{~Rahman et al.}

\institute{Ca' Foscari University of Venice
\email{\{muhammad.rahman,sebastiano.vascon\}@unive.it}\and
University of Padova \email{\{piero.simonetto,anna.polato\}@phd.unipd.it}, \email{\{francesco.pasti\}@dei.unipd.it},\email{\{luca.tonin\}@unipd.it}\and
European Centre for Living Technology}

\maketitle

\begin{abstract}
Open vocabulary 3D object detection (OV3D) allows precise and extensible object recognition crucial for adapting to diverse environments encountered in assistive robotics. This paper presents \model, a zero-shot 3D object detection pipeline based on RGB-D images for smart wheelchairs. Our pipeline integrates an open-vocabulary 2D object detector with a mask generator for semantic segmentation, followed by depth isolation and point cloud construction to create 3D bounding boxes. The smart wheelchair exploits these 3D bounding boxes to identify potential targets and navigate safely. We demonstrate \model's performance through experiments on the Replica dataset and we report preliminary results with a real wheelchair. \model~improves state-of-the-art significantly on the Replica dataset at mAP25 (+9pts) and mAP50 (+5pts) with marginal improvement at mAP. The code is publicly available at this link \href{https://github.com/EasyWalk-PRIN/OpenNav}{https://github.com/EasyWalk-PRIN/OpenNav}.
\end{abstract}

\section{Introduction}
\begin{figure}[t]
    \centering
    \includegraphics[width=0.6\textwidth]{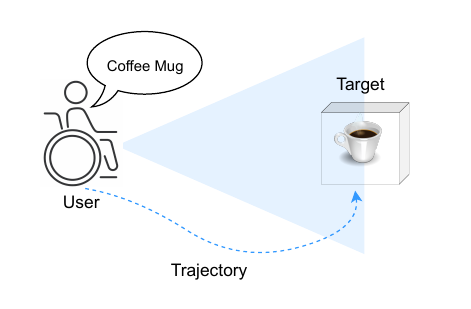} 
    \caption{A wheelchair user inputs a desired object through a prompt. A sophisticated 3D object recognition system then scans the environment to locate the specified object. Once identified, the wheelchair navigates smoothly towards the chosen target.}
    \label{fig:teaser} 
\end{figure}

The integration of advanced computer vision and robotics has the potential to greatly improve smart wheelchair navigation for individuals with mobility impairments~\cite{808948, 7759291}. Developing robust navigation systems for different and unknown environments is a critical challenge, requiring sophisticated perception capabilities, particularly in precise and adaptable target object detection. Conventional navigation methods often struggle in new or evolving environments and may depend on predetermined maps or limited identifiable objects~\cite{5625092}. Accurate 3D data is essential for secure and effective navigation, especially in wheelchair mobility. Through a sophisticated 3D object recognition system, a wheelchair user can specify a desired object name to locate the object in the environment and navigate towards it, enhancing autonomy and safety, as illustrated in Figure~\ref{fig:teaser}.

Recent advancements in computer vision, particularly in open-vocabulary and zero-shot learning domains~\cite{clip}, offer promising possibilities to overcome these limitations~\cite{homerobot,liu2024mokaopenvocabularyroboticmanipulation}. Open-vocabulary object detection allows for recognizing a wide range of objects without needing extensive training on specific categories~\cite{yoloworld}. This flexibility is essential for assistive robots to adapt to diverse environments and user needs. Zero-shot learning expands this capability further, enabling the detection of previously unseen objects based on semantic understanding~\cite{yoloworld}. 

Robotic navigation approaches can be broadly categorized into map-based methods \cite{eng4020092} or target-driven navigation~\cite{10341666}. Map-based methods are effective in stable environments but struggle with unknown scenarios and require extensive pre-mapping. Target-driven navigation, closely aligned with the needs of assistive robotics, has historically been limited by the scope of detectable objects and the accuracy of 3D localization~\cite{10323166}.

Herein, we propose \model, a zero-shot 3D object detection and navigation pipeline designed for wheelchair navigation. \model~ uses RGB-D images from an RGB-D camera and processes the data through an open-vocabulary 2D object detector, a mask generator for semantic segmentation, depth isolation, and point cloud reconstruction. The resulting 3D bounding boxes enable the wheelchair to identify targets and navigate safely around obstacles. This integration allows for precise and extensible object recognition, which is crucial for adapting to a variety of daily life scenarios. The open-vocabulary nature of our system means it can be easily extended to recognize new objects or adapt to specific user needs without requiring extensive retraining.
Our key contributions of this paper are:
\begin{enumerate}
    \item An open vocabulary zero-shot 3D object detection pipeline capable of recognizing and localizing novel objects in real-world environments for robotic navigation.
    \item Experimental validation of \model~on the Replica dataset reporting state-of-the-art performance.
    \item Real-world tests with a smart wheelchair equipped with our system demonstrate its effectiveness in 3D object detection and accurate target identification.
\end{enumerate}

\section{Related Work}
\subsection{Object Detection for Navigation}
Recent advances in object detection for robot navigation have significantly improved the capabilities of robotic systems ~\cite{9796979,homerobot,liu2024mokaopenvocabularyroboticmanipulation}. For example, a lightweight method has been proposed that combines multiple efficient detectors to achieve high-accuracy 3D dynamic obstacle detection, which is crucial for robots with limited computational resources ~\cite{10323166}. By fusing 2D laser and RGB camera information, the detection of a wide range of classes is enabled, enhancing object localization and navigation in complex environments ~\cite{9837249}. A few-shot object detection method has also been proposed, allowing robots to learn new objects with minimal annotation, which is valuable for adaptive navigation scenarios ~\cite{10030974}. Additionally, the adaptation of YOLOv3\cite{redmon2018yolov3} for detecting and tracking specific objects such as doors improves the autonomy and navigation precision of smart wheelchairs ~\cite{Lecrosnier2020DeepLO}. An egocentric computer vision system for robotic wheelchairs has been proposed, enabling hands-free motion control, autonomous navigation to various targets, and adaptive learning based on user preferences ~\cite{ego}. In this paper, we propose a 3D object detection system for intelligent wheelchairs, leveraging zero-shot learning for open-vocabulary recognition, offering more flexibility in detection.

\subsection{Open Vocabulary Object Detection}
Open vocabulary object detection allows for detecting objects that belong to known classes and those not seen during training based on a text description~\cite{yoloworld, gdino, ovdetr, openscene2d}. These methods typically involve aligning image and text features to generate bounding boxes corresponding to the most relevant text labels. Various approaches have been proposed for open vocabulary 3D object detection~\cite{ov3det, fmov3d, coda}, which align textual and point cloud representations for novel class discovery. Several approaches are proposed for 3d instance segmentation~\cite{openmask3d, openScene, huang2023openins3d}. OVIR3D~\cite{ovir3d} produces a 3D instance mask by fusing text-aligned 2D proposals from a 2D instance segmentation model. 
Open3DIS~\cite{open3dis} combines 3D instance networks with a 2D-guided 3D Instance Proposal Module, utilizing agglomerative clustering to fuse 2D masks, and processes the resulting high-quality 3D object proposals with pointwise 3D CLIP features for open-vocabulary instance segmentation. 
Our \model~differentiates from the others by relying solely on a real-time open vocabulary 2D object detector and 2D mask generator without the need of storing nor generating the whole point cloud of the scene (apart for evaluation purposes). We reconstruct the segmented point cloud of each object using camera intrinsic parameters and create a 3D bounding box for each object. This process allows for 3D instance segmentation-based object detection.

\section{Proposed pipeline}

\begin{figure}[t!]
    \centering    \includegraphics[width=1.0\textwidth]{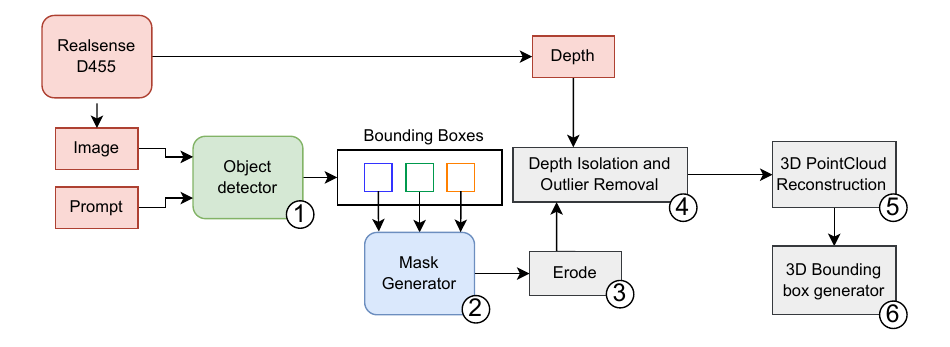} 
    \caption{Open vocabulary 3D Object detection and reconstruction pipeline. Our pipeline works by using input from RGB-D camera. RGB image and textual description of desired class is fed into 2D object detector to generate 2D bounding boxes. Then, the mask generator generates a semantic mask for objects in bounding boxes. The masks are filtered, and then the depth is isolated based on masks to reconstruct each object's point cloud. Lastly, we create 3D bounding boxes around the point cloud of each object.}
    \label{fig:wheelchair}
\end{figure}

\begin{figure}[t]
    \centering
    \includegraphics[width=0.9\textwidth]{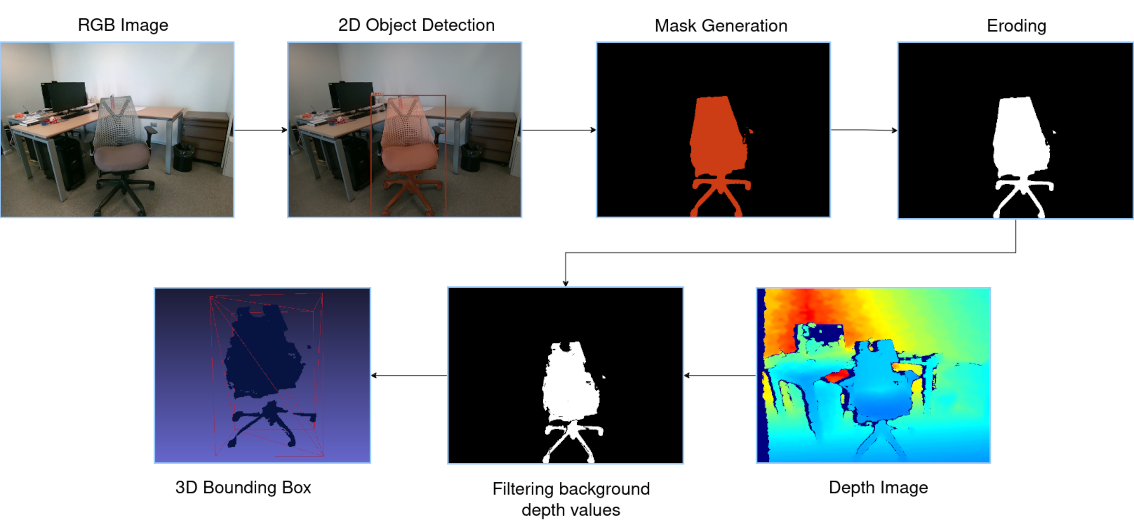} 
    \caption{From RGBD to 3D bounding box generation using input from Realsene camera.}
    \label{fig:3dpointcloud} 
\end{figure}

We capture RGB and depth data using an Intel RealSense D455 camera. The RGB image $\mathcal{I}_{rgb} \in \mathbb{R}^{H \times W \times 3}$ and depth image $\mathcal{I}_{d} \in \mathbb{R}^{H \times W}$ are synchronized to ensure consistency and alignment between the two modalities. We also have access to the camera's intrinsic and extrinsic parameters.
Our model is composed of a pipeline of several (six) steps (see Figure~\ref{fig:wheelchair}). Each step is marked with a number (e.g. \circled{1}...\circled{6}) reported in the corresponding subsection. In figure \ref{fig:3dpointcloud} we show qualitatively each step. We also sketched our method's pseudocode in Alg.~\ref{alg:algo}.

\subsubsection{2D object detection}\circled{1} 
We utilize an open vocabulary 2D object detector to perform object detection on RGB image $\mathcal{I}_{rgb}$. Given the input image and text $\mathcal{T}_{C}$ consisting of classes of interest $\mathcal{C}$, the 2D object detector outputs bounding boxes $\mathcal{B}$, scores $\mathcal{S}$ and class labels $\mathcal{L}$. These bounding boxes are used to localize objects in the image. 
Given $\mathcal{I}_{rgb}$ and $\mathcal{T}_{C}$, the output of the object detector is:
\begin{equation}
\{(b_i, s_i, l_i) \mid i = 1, \ldots, k\} = \text{ObjectDetector}(\mathcal{I}_{rgb}, \mathcal{T}_{C})
\end{equation}
Where:
\begin{itemize}
    \item $b_i = (x_{i1}, y_{i1}, x_{i2}, y_{i2})$ represents the bounding box coordinates (top-left corner $(x_{i1}, y_{i1})$, bottom-right corner $(x_{i2},y_{i2})$) of the $i_{th}$ bounding box,
    \item $s_i \in [0, 1]$ is the confidence score for the detection,
    \item $l_i \in \mathcal{C}$ is the class label for the detected object.
\end{itemize}
\subsubsection{Mask Generator}\circled{2} 
Given an RGB image $\mathcal{I}_{rgb}$ and a set of bounding boxes $b_i$, where each bounding box constitutes a potential object in the image, our goal is to generate a corresponding set of masks $\mathcal{M} = \{m_1, m_2, ..., m_i\}$. As we know the labels of bounding box, each mask $m_i \in \{0, 1\}^{H \times W}$ indicates the pixels belonging to the object. We utilize a mask generator to output the mask which is given as:
\begin{equation}
\ m_i = \text{MaskGenerator}(\mathcal{I}_{rgb}, b_i)
\end{equation}
Where
\begin{equation}
m_i(u, v) = \begin{cases} 
1 & \text{if pixel } (u, v) \text{ belongs to the object in } b_i \\
0 & \text{otherwise}
\end{cases}
\end{equation}
% \subsubsection{PostProcessing}
\subsubsection{Morphological Erosion} \circled{3}  We apply morphological erosion to refine the masks $\mathcal{M}$ and remove background noise. Let $\mathcal{M}' = \{m_1', m_2', \ldots, m_k'\}$ be the set of eroded masks. Mathematically, erosion is defined as:
\begin{equation}
m_i' = m_i \ominus \mathcal{K}
\end{equation}

Where $\ominus$ denotes the morphological erosion operation and $\mathcal{K}$ is a structuring element ($3\times3$ matrix).

\subsubsection{Depth Isolation and Z-Score Filtering} \circled{4} Using the eroded masks $\mathcal{M}'$, we isolate the depth values from $\mathcal{I}_{d}$ corresponding to each object. For each mask $m_i'$, the isolated depth map $\mathcal{D}_i$ is given by:
\[
\mathcal{D}_i(u, v) = \begin{cases} 
\mathcal{I}_{d}(u, v) & \text{if } m_i'(u, v) = 1 \\
0 & \text{otherwise}
\end{cases}
\]

Next, we apply Z-score based filtering to remove depth outliers. For each $\mathcal{D}_i$, we compute the mean $\mu_i$ and standard deviation $\sigma_i$ of the non-zero depth values. We then filter the depth values using the Z-score:
\[
\mathcal{D}_i'(u, v) = \begin{cases} 
\mathcal{D}_i(u, v) & \text{if } \left| \frac{\mathcal{D}_i(u, v) - \mu_i}{\sigma_i} \right| < \tau \\
0 & \text{otherwise}
\end{cases}
\]
where $\tau$ is a predefined Z-score threshold set equal to $2$. The resulting depth-filtered masks are denoted as $\mathcal{M}'' = \{\mathcal{D}_1', \mathcal{D}_2', \ldots, \mathcal{D}_i'\}$.
\subsubsection{Object Point-Cloud Reconstruction}\circled{5} 
The 3D point cloud reconstruction process transforms the 2D depth information from the RGBD data into a 3D spatial representation. This transformation requires the intrinsic and extrinsic parameters of the camera, and we gather them from the IntelRealSense D455. We reconstruct the aligned point cloud for each detected object using its isolated depth.

\paragraph{Intrinsic Parameters}

The intrinsic parameters of a camera define the mapping between the 3D scene and the 2D image plane. These parameters include the focal lengths and the principal point offsets, which are usually represented in a camera calibration matrix $\mathbf{K}$:

\begin{equation}
\mathbf{K} = \begin{pmatrix}
f_x & 0 & c_x \\
0 & f_y & c_y \\
0 & 0 & 1
\end{pmatrix}
\end{equation}

where:
\begin{itemize}
    \item $f_x$ and $f_y$ are the focal lengths in the x and y directions, respectively.
    \item $(c_x, c_y)$ is the principal point, i.e., the coordinates of the optical center in the image plane.
\end{itemize}

Given an object $i$ and its depth values $\mathcal{D}'_i$, the 3D coordinates $(X, Y, Z)$ of each pixel $(u, v)$  can be computed as:
\begin{align*}
X &= (u - c_x) \cdot \frac{\mathcal{D}'_i}{f_x}           &  Y &= (v - c_y) \cdot \frac{\mathcal{D}'_i}{f_y}              &  Z &= \mathcal{D}'_i(u, v)
\end{align*}

This transformation maps each pixel in the 2D depth image to a point in the 3D space.

\paragraph{Extrinsic Parameters}

The extrinsic parameters define the transformation from the camera's coordinate system to the world coordinate system. These parameters are represented by a rotation matrix $\mathbf{R}$ and a translation vector $\mathbf{t}$, which together form the extrinsic transformation matrix $\mathbf{T}$:

\begin{equation}
\mathbf{T} = \begin{pmatrix}
\mathbf{R} & \mathbf{t} \\
0 & 1
\end{pmatrix}
\end{equation}

where:
\begin{itemize}
    \item $\mathbf{R}$ is a $3 \times 3$ rotation matrix.
    \item $\mathbf{t}$ is a $3 \times 1$ translation vector.
\end{itemize}

The complete transformation from the camera coordinate system to the world coordinate system for a point $(X_c, Y_c, Z_c)$ in the camera frame is given by:
\begin{equation}
\begin{pmatrix}
X_w \\
Y_w \\
Z_w \\
1
\end{pmatrix}
=
\mathbf{T}
\begin{pmatrix}
X_c \\
Y_c \\
Z_c \\
1
\end{pmatrix}
=
\begin{pmatrix}
\mathbf{R} & \mathbf{t} \\
0 & 1
\end{pmatrix}
\begin{pmatrix}
X_c \\
Y_c \\
Z_c \\
1
\end{pmatrix}
\label{eqn:worldcoordinate}
\end{equation}

\subsubsection{3D Bounding Boxes Generation}\circled{6} 
%BBOX thickness from prior or from LLM \sv{NEED TO WRITE THIS!}
The 3D bounding boxes are generated from each 3D instance segmentation. In particular, once $X_w$, $Y_w$, $Z_w$ (see Eqn.\ref{eqn:worldcoordinate})  are computed for an object, their extreme values ($\min$ and $\max$ in each dimension) represent the bounding box corners. We opted for this simple strategy because it was fast to compute and functional for navigation purposes. Indeed, more sophisticated methods might be used here, for example, apriori knowledge of each object size can be used.\\
~\\
The 3D bounding boxes of the objects are then used as targets for robotic navigation.

\input{alg}

\section{Experiments}
\subsection{Dataset}
We conducted experiments using the Replica dataset~\cite{replica}, a high-fidelity synthetic dataset comprising 48 distinct classes. Replica provides $8$ evaluation scenes to evaluate the performance of 3D instance segmentation. As our 3D bounding box generation relies on the segmentation of objects, we use the Replica dataset to evaluate the performance in instance segmentation.
To complement our synthetic dataset experiments, we extend our application to real-world scenarios of assistive wheelchairs. In particular, we use the \model~for target detection for semi-autonomous wheelchair navigation. This study enables the user to specify any desired object through a text prompt, and the wheelchair automatically navigates toward it. 

\subsection{Implementation Details}
\model~is built on top of a generic open vocabulary 2D object detector and a mask generator. \model~do not require any fine-tuning, nor ad-hoc training, since we leverage only pretrained models. Our baseline pipeline is built on top of YoloWorld-S~\cite{yoloworld} for 2D object detection and MobileSAM~\cite{mobile_sam} for object segmentation. The users can choose the object detection model from YoloWorld-S, YoloWorld-M and YoloWorld-L.  The more advanced version of our pipeline includes a combination of Grounding-DINO Swin-T~\cite{gdino} for the 2D object detection and RepViT-SAM~\cite{repvitsam} for the segmentation parts, respectively. We use the Scipy~\cite{scipy} python library for z-score calculation and Open3D~\cite{open3d} for 3D bounding box generation and point cloud visualizations. The experiments are conducted on Quadro RTX 5000 16GB RAM. Furthermore, the whole 3D object detection pipeline is implemented in ROS2 Humble~\cite{ros2} to be easily interfaced with the robotic community.

\subsection{Metrics and Competitors}

\subsubsection{Competitors} We assess  \model's performance against five state-of-the-art competitors, OVIR-3D~\cite{ovir3d},       Open3DIS~\cite{open3dis},
OpenIns3D~\cite{huang2023openins3d}, OpenScene~\cite{openScene}, OpenMask3D~\cite{openmask3d} on the Replica dataset. These are all open vocabulary models that leverage object detectors and segmentation methods to locate objects in 3D, as per our pipeline.

For the sake of fairness, our top pipeline uses a more recent segmentation backbone than the one proposed by our competitors (in particular Open3DIS~\cite{open3dis} that shares our detection module, GDINO). 
Since our model considers one viewpoint at a time (one Replica's scene is composed of 200 viewpoints), we merged instances of the same class if their IoU is greater than 0.8 and then assessed the quantitative performance.

\subsubsection{Evaluation Criteria} We followed the same evaluation criteria as in~\cite{open3dis}. Therefore, we report performance considering standard mean Average Precision (mAP) metric and mAP at Intersection over Union thresholds of 50\% (mAP50) and 25\% (mAP25). 

\subsection{Results}
\begin{figure}[h!]
    \centering
\includegraphics[width=1\textwidth]{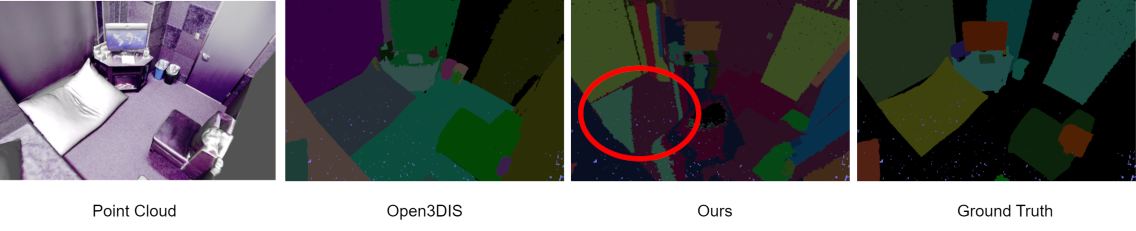} 
    \caption{Failure case. In the case of large objects that are not entirely visible in a single view, they might not be well detected. This happens, for example for the pillow (red circle). Therefore, such objects get broken into different segments in the subsequent phases resulting in different instances.}
    \label{fig:fcase}
\end{figure}

We reported the mAPs results on Replica dataset in Table \ref{tab:sota_replica}, from which we can draw some considerations. In general, the proposed method considerably outperformed all the competitors when the IoU is relaxed (from 50\% to 25\%). At the same time, we perform slightly better than the best competitor (Open3DIS~\cite{open3dis}) when no IoU threshold is set. This is explained by the fact that our method is not exploiting the whole point cloud (as~\cite{open3dis} and other does) of the scene on inference or when doing performance evaluation; therefore, certain classes that occupy a vast part of the scene (like desks, beds, tables, tv-screen, sofa,...) cannot be fully detected in one view, hence requires a relaxed IoU (see Figure \ref{fig:fcase} in which the pillow object is broken into different instances). The second consideration is that the better the detector and segmentation model, the higher the performance. This is clear from the results where our best model, equipped with strong detector and segmentation models (we refer to "GDino+RepViTSAM (ours)"), achieves better performance than others that rely on much more complex architectures and/or more information (e.g., the full point cloud of the scene).

\begin{table}[t]
  \centering
  \caption{State-of-the-art comparison on Replica dataset.}
  \label{tab:sota_replica}
  
  \begin{tabular}{*{4}{c}}
    \toprule
      \textbf{Method} & \textbf{mAP} & \textbf{mAP50} & \textbf{mAP25} \\
    \midrule
      OVIR-3D (2D fusion) ~\cite{ovir3d} & 11.1  & 20.5 & 27.5  \\
      OpenScene (2D fusion) ~\cite{openScene}  & 10.9 & 15.6 & 17.3 \\
      OpenScene (3D Distill)~\cite{openScene} &  8.2 & 10.5  & 12.6 \\
      OpenScene (2D-3D)~\cite{openScene} &  8.2 &  10.4 & 13.3 \\
      OpenMask3D~\cite{openmask3d} & 13.1  & 18.4  & 24.2  \\
      Open3DIS (3D) ~\cite{open3dis} & \underline{18.5}  & \underline{24.5} & \underline{28.2} \\

      Open3DIS (2D-3D) ~\cite{open3dis} & 14.9 & 18.8 &  23.6 \\ 
      OpenIns3D (3D) ~\cite{huang2023openins3d} & 13.6 & 18.0 & 19.7 \\
    \midrule
     \model: YoloWorld+MobileSAM (ours) (2D)  & 8.5 & 15.1 & 20.4 \\
     \model: GDino+RepViTSAM (ours) (2D) & \textbf{18.7} & \textbf{29.7} & \textbf{37.7}\\
    \bottomrule
  \end{tabular}
\end{table}

\subsubsection{Running time evaluation} Running time is also a crucial aspect when deploying computer vision systems into robots (like in our assistive wheelchair setup) that interact with the environment. In table \ref{tab:running_time}, we report the time in seconds needed to process each scene compared with our main competitor, Open3DIS. As can be seen, our method drastically reduced the inference time, achieving a speedup of $\times$2-4 based on the different backbone models.
\begin{table}[h!]
\centering
  \caption{Running time of our models and Open3DIS. We reported the time per scene and the average time per view considering 200 views per scene.}
  \label{tab:running_time}
\begin{tabular}{@{}lcc@{}}
\toprule
\textbf{Method} & \textbf{Secs per scene} & \textbf{Secs per view}  \\ \midrule
Open3DIS & 197.32 & 0.98 \\
\midrule 
Our (YoloWorld+MobileSAM) & 27.76 & 0.13 \\
Our (GDino+RepViTSAM) & 94.35 & 0.47 \\ \bottomrule
\end{tabular}
\end{table}
\subsection{Ablation}
We performed an ablation study considering different pipeline combinations by changing detection and segmentation models. The different combinations are evaluated on the Replica dataset, and the results are reported in Table \ref{tab:ablation}. We found that the main gain is due to the detection system rather than the segmentation models. Better segmentation models only bring a small improvement of a few decimal points, while different detection modules double the performance.
\begin{table}[t]
  \centering
  \caption{Ablation study on different combinations of detector and segmentation models. The main gain is due to the GDino VS YoloWorld accuracies. While changing segmentation models, the gain is negligible.}
  \label{tab:ablation}
  
  \begin{tabular}{*{4}{c}}
    \toprule
      \textbf{Method} & \textbf{mAP} & \textbf{mAP50} & \textbf{mAP25} \\
     \model: YoloWorld+MobileSAM (ours)   & 8.5 & 15.1 & 20.4 \\
     \model: YoloWorld+RepVitSAM (ours)   & 9.8 & 15.9 & 20.6 \\
     \model: GDino+MobileSAM (ours) & 18.3 & 29.3 & 37.4\\
     \model: GDino+RepViTSAM (ours) & \textbf{18.7} & \textbf{29.7} & \textbf{37.7}\\

     \bottomrule
  \end{tabular}
\end{table}

\subsection{Qualitative analysis}
\begin{figure}[h]
    \centering    \includegraphics[width=1.0\textwidth]{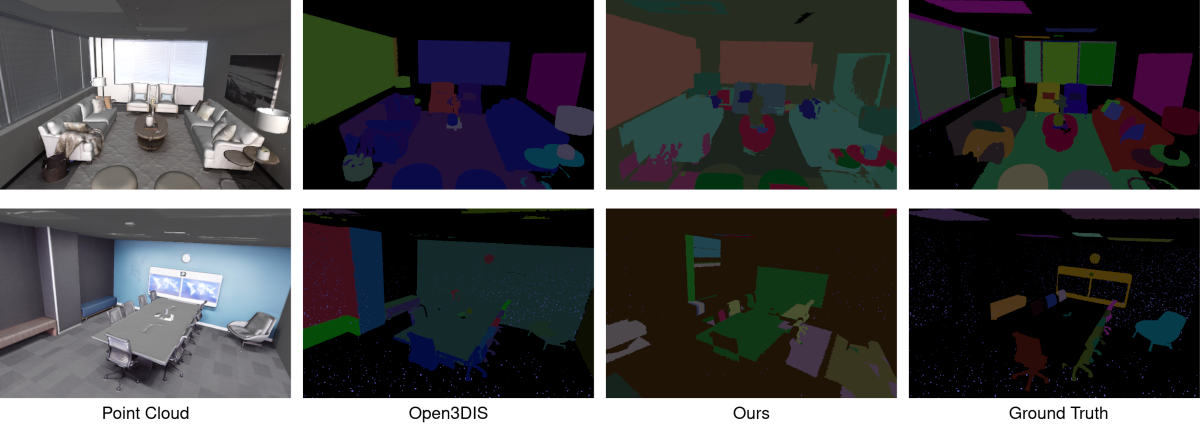} 
    \caption{Instance segmentation results on Replica}
    \label{fig:qual}
\end{figure}

In Figure \ref{fig:qual} we reported some qualitative results. We compared our best model with Open3DIS. As can be seen, in the first row, our model is able to segment the small table in the center of the room, the whole sofa, and part of the pillow in the rightmost sofa, while breaking the window into multiple instances (an object that Open3DIS can capture entirely). Considering the second row, we were able to capture the TV screen without confusing it with the wall, but we failed to discern the central chair and the table. This artifact happens due to the masking of the depth that produces shadows that are reflected in the segmentation.

\section{Case study on a smart powered wheelchair}
We evaluated the proposed \model~algorithm for the indoor navigation of a smart powered wheelchair. The goal is to demonstrate that the algorithm is able: i) to quickly detect the required objects in the surroundings; ii) to reliably track the detected object during wheelchair motion; and iii) to provide an accurate estimation of the object's position.

\subsection{The smart wheelchair setup}
\begin{figure}[t]
\centering
\includegraphics[width=1.0\textwidth,trim=0.1 0.1 0.1 0, clip]{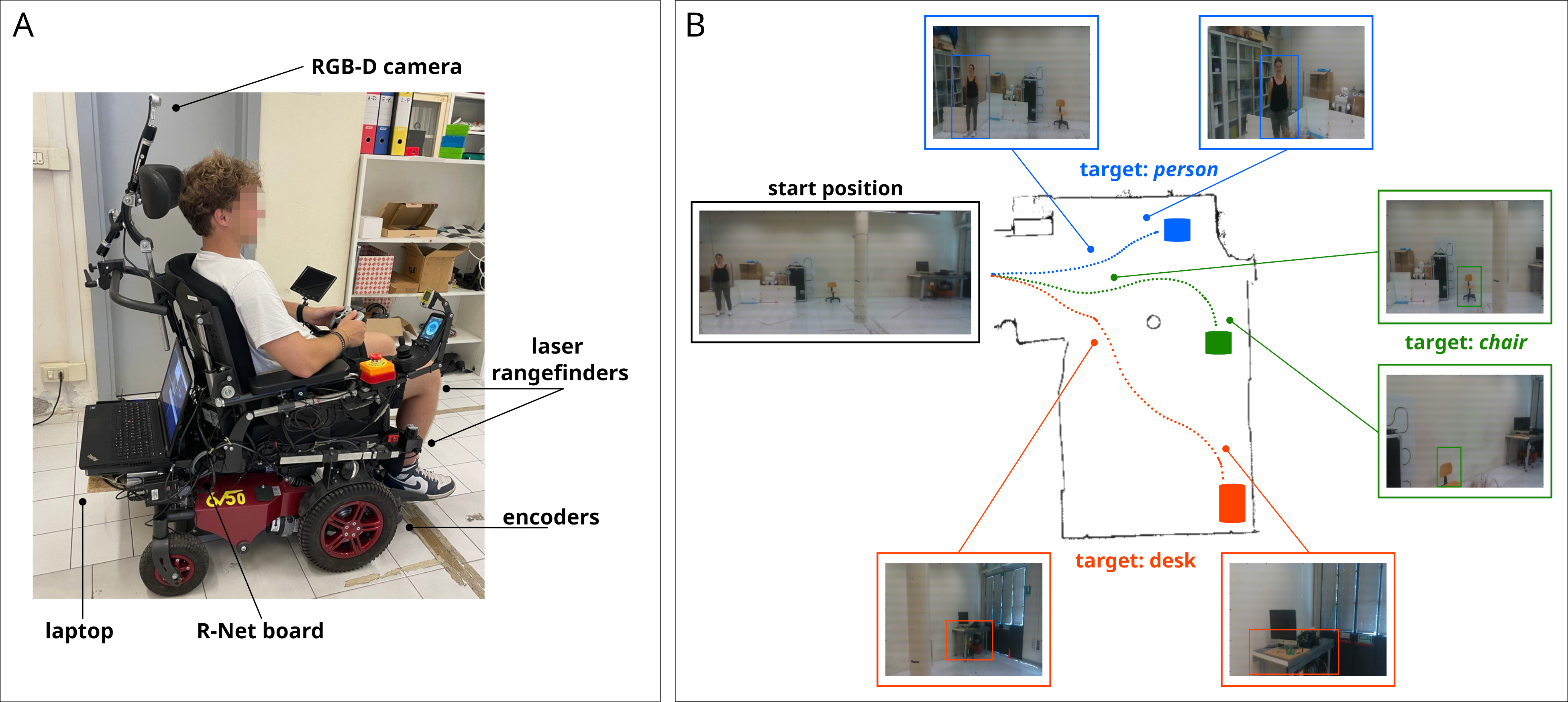}
\caption{A) Side-view of the wheelchair setup with the equipped sensors. Encoder and laser rangefinder are duplicated and symmetrically placed on the other side of the wheelchair. B) Experimental results of the wheelchair navigation with three targets (e.g., person, chair, and desk, in blue, green, and orange, respectively) provided by the \model~algorithm. Dotted lines show the real trajectories performed by the wheelchair to reach each target. Image in the boxes show the target detection (e.g., bounding box) during the motion and at the end position of the wheelchair.}
\label{fig:wheelchair_results}
\end{figure}

The powered wheelchair (HI-LO Vario) was a commercial device provided by the company Vassilli s.r.l., Padua, Italy (Figure~\ref{fig:wheelchair_results}A). We equipped the wheelchair with two custom-made incremental encoders placed on the actuated wheels. A simple two-pulley design was used to translate the encoder’s shaft rotation into wheel rotations. The encoder ticks were acquired using an Arduino Mega and then published in the~\ac{ros} middleware, where they were used to compute wheel odometry by following the principles of differential drive odometry kinematics~\cite{xu2019}.

For optimal environment awareness, a variety of detection technologies were incorporated into the design. Two laser rangefinders (URG-04LX-UG01, Hokuyo Automatic Co LTD, Japan) were placed on the front right and left, offering a complete view of the setting surrounding the frontal part of the wheelchair and covering most of the lateral sides. An RGB-D camera (Intel RealSense D435, Intel, US) was placed on top of the wheelchair with a metallic extension arm to provide 3D point clouds, depth maps, and color perception information, which were used by the \model~algorithm.

Finally, an R-Net board was used to interface with the CAN-bus of the wheelchair and control the motors through a USB connection with the laptop placed on the back.

\subsection{The semi-autonomous navigation system}
The navigation system of the wheelchair was implemented in \ac{ros} and was based on the integration of \ac{apf} and an \ac{mpc} controller~\cite{messiou2022, delazzari2024}. Briefly, two dynamic equations were controlling the wheelchair's motion for the angular and linear velocity, respectively~\cite{tonin2022}. On the one hand, information on obstacles in the environment was gathered by the laser rangefinders and converted into repellers for the \ac{apf} algorithm. The resulting force was used to determine the angular velocity of the wheelchair. On the other hand, the linear velocity of the wheelchair was inversely proportional to the distance of the closest detected obstacle. It has been demonstrated that such a navigation system safely avoids obstacles, generates smooth trajectories, and provides a natural behaviour of the robotic device~\cite{beraldo2022, tonin2022}.

The semi-autonomous behavior of the wheelchair is implemented by converting user intentions into virtual attractors for the navigation system. Then, \ac{mpc} acted as a local planner to generate the proper trajectory for the wheelchair. The combination of the \ac{apf} and \ac{mpc} controls the angular velocity of the wheelchair by rotating it towards the desired directions~\cite{messiou2022, delazzari2024}. In general, user intentions can be provided by a variety of \ac{at} interfaces ranging from head-switches and mouth-pads to more sophisticated systems such as \acp{bmi}~\cite{beraldo2022, tonin2022, tortora2022}. In this work, targets (i.e., attractors for the \ac{apf}) were provided by the \model~algorithm.

\subsection{Preliminary navigation results}
Figure~\ref{fig:wheelchair_results}B shows the experimental results of the wheelchair's semi-autonomous navigation system driven by the \model~algorithm. The target (person, chair, or desk) was provided to the system at the beginning of each run. Then, the wheelchair moved from the start position towards the corresponding target. In the case of the first target, the wheelchair had to pass the bookshelf on the left and move towards the \textit{person} (in blue in the figure). For the \textit{chair} (in green), the wheelchair had to avoid the column in the middle of the room and reach the target. Finally, for the last target (\textit{desk}, in orange), the wheelchair had to avoid the column and move to the right part of the room. The dotted lines in the figure show the trajectories for each target computed by exploiting the odometry of the wheelchair. Furthermore, the images in the boxes depict the detected objects (bounding boxes) during the motion and at the end of the path.

\section{Conclusions \& Limitations}
In this paper, we propose \model, an open-vocabulary 3D object detection pipeline that can be integrated into any robotic system to help in navigation. We developed our model in ROS2 so that it can be easily integrated into existing robotic pipelines. We quantitatively demonstrate on the Replica dataset that our straightforward method, leveraging recent pre-trained open vocabulary models, surpasses state-of-the-art methods trained on specific datasets. Furthermore, we tested the system in a real robotic environment, showcasing its feasibility for real-world assistive robotic applications.\\
~\\
As for the \emph{limitations}, our model performance is upper-bounded by the segmentation and detection modules, therefore, combining poor/strong detection or segmentation systems we lead to poorer/outstanding results. Concerning the assistive robotic application, we assumed at this stage that there is only one object per class in the scene and that the user is looking for only one object. Multiple instances of the same object class will be investigated in the future.\\
~\\
Through \model, we demonstrate the potential of advanced computer vision techniques and pre-trained models, to improve assistive technologies significantly. 

\section*{Acknowledge}
This study was carried out within the project PRINPNRR22 “EasyWalk: Intelligent Social Walker for active living” and received funding from the European Union Next-GenerationEU - National Recovery and Resilience Plan (NRRP) – MISSION 4 COMPONENT 2, INVESTIMENT 1.1 – CUP N. H53D23008220001. This manuscript reflects only the authors’ views and opinions, neither the European Union nor the European Commission can be considered responsible for them. We want to thank Dr. Luca Palmieri for the helpful discussions.

% \clearpage\mbox{}Page \thepage\ of the manuscript.
% \clearpage\mbox{}Page \thepage\ of the manuscript.
% \clearpage\mbox{}Page \thepage\ of the manuscript.
% \clearpage\mbox{}Page \thepage\ of the manuscript.
% \clearpage\mbox{}Page \thepage\ of the manuscript. This is the last page.
% Now we have reached the maximum length of an ECCV \ECCVyear{} submission (excluding references).
% References should start immediately after the main text, but can continue past p.\ 14 if needed.
% \clearpage  % TODO REVIEW/FINAL: This \clearpage needs to be removed from both review and camera-ready versions.

\bibliographystyle{splncs04}
\bibliography{main}
\end{document}

%% file: alg.tex
\begin{algorithm}[h]
  \SetAlgoLined
  \SetKwInOut{Input}{Input}
  \SetKwInOut{Output}{Output}

  \Input{RGB image \( I_{rgb} \), Depth map \( I_{d} \), Classes \( C = \{c1, c2, ..., cn\} \)}
  \Output{3D bounding boxes around detected objects}

  \BlankLine
  \textbf{Step 1: Object Detection}\;
  \( \mathcal{B}, \mathcal{S}, \mathcal{L} \) = ObjectDetector(\( I_{rgb}, \mathcal{C} \))\;

  \BlankLine
  \textbf{Step 2: Mask Generation}\;
  \( M \) = MaskGenerator(\( I_{rgb}, \mathcal{B} \))\;

  \BlankLine
  \textbf{Step 3: Morphological Erosion}\;
  \( M' \) = morphologicalErosion(\( M \))\;

  \BlankLine
  \textbf{Step 4a: Depth Isolation}\;
  Extract depth values from \( I_{d} \) using \( M' \)\;

  \BlankLine
  \textbf{Step 4b: Z-score Filtering}\;
  Apply Z-score filtering on \(M'\) to remove outliers to get \( M'' \) \;

  \BlankLine
  \textbf{Step 5: 3D Object Point Cloud Reconstruction}\;
  Reconstruct 3D point clouds using \( M'' \) and depth data\;

  \BlankLine
  \textbf{Step 6: 3D Bounding Box Generation}\;
  Generate 3D bounding boxes around each object\;

  \caption{3D Object Detection and Reconstruction Algorithm}
  \label{alg:algo}
\end{algorithm}